\documentclass{article}

\usepackage{lmodern}
\usepackage[utf8]{inputenc}
\usepackage[T1]{fontenc}
\usepackage[english]{babel}
\usepackage{csquotes}

\usepackage{amsmath,amssymb,amsthm}
\usepackage{mathtools,thmtools,stmaryrd,esint}
\usepackage{mathrsfs}

\usepackage{hyperref}
\hypersetup{colorlinks=true}
\usepackage[shortlabels]{enumitem}
\usepackage[nameinlink]{cleveref}
\crefname{equation}{}{}
\crefname{enumi}{}{}

\usepackage[svgnames]{xcolor}
\usepackage{graphicx}
\usepackage{todonotes}
\usepackage{longtable,tabu}
\usepackage[shortlabels]{enumitem}
\usepackage{bm} 

\numberwithin{equation}{section}

\declaretheorem[name=Theorem,within=section]{thm}

\declaretheorem[name=Proposition,numberlike=thm]{prp}

\declaretheorem[name=Remark,numberlike=thm,style=remark]{rmk}

\declaretheorem[name=Example,numberlike=thm,style=remark]{ex}

\usepackage{microtype}
\usepackage{graphicx}
\usepackage{subfigure}
\usepackage{booktabs} 
\usepackage{enumitem} 

\usepackage[accepted]{icml2024}

\icmltitlerunning{Generalized grammar rules and structure-based generalization beyond classical equivariance}

\begin{document}

\twocolumn[
\icmltitle{Position Paper: Generalized grammar rules and structure-based generalization beyond classical equivariance for lexical tasks and transduction}



\begin{icmlauthorlist}
\icmlauthor{Mircea Petrache}{mp}
\icmlauthor{Shubhendu Trivedi}{st}
\end{icmlauthorlist}

\icmlaffiliation{mp}{Department of Mathematics and Institute for Math. and Comp. Engineering, PUC Chile}
\icmlaffiliation{st}{shubhendu@csail.mit.edu}

\icmlcorrespondingauthor{Mircea Petrache}{mpetrache@mat.uc.cl}
\icmlcorrespondingauthor{Shubhendu Trivedi}{shubhendu@csail.mit.edu}

\icmlkeywords{Machine Learning, ICML}

\vskip 0.3in
]



\printAffiliationsAndNotice{} 

\begin{abstract}
Compositional generalization is one of the main properties which differentiates lexical learning in humans from state-of-art neural networks.
We propose a general framework for building models that can generalize compositionally using the concept of Generalized Grammar Rules (GGRs), a class of symmetry-based compositional constraints for transduction tasks, which we view as a transduction analogue of equivariance constraints in physics-inspired tasks.
Besides formalizing generalized notions of symmetry for language transduction, our framework is general enough to contain many existing works as special cases.
We present ideas on how GGRs might be implemented, and in the process draw connections to reinforcement learning and other areas of research.
\end{abstract}

\section{Introduction}

Humans are able to effortlessly process sentences that they have never encountered before. This is true not just for meaningful sentences which contain words whose meanings are not known, but also for nonsense sentences peppered with nonce words. Such sentences are semantically meaningless, but can still be evocative from their syntax. Consider the example provided by \cite{Ingraham1903}: \emph{the gostak distims the doshes.} On the surface, the sentence is meaningless, but as \cite{Ingraham1903} points out: 
``You do not know what this means; nor do I. But if we assume that it is English, we know that the \emph{doshes are distimmed by the gostak}. We know too that \emph{one distimmer of doshes is a gostak}. If, moreover, the doshes are galloons, we know that \emph{some galloons are distimmed by the gostak}. And so we may go on.''


This remarkable cognitive ability is a result of \textbf{compositional generalization}~\cite{Chomsky1957, Montague1970, fodor1993connectionism, Partee1995LexicalSA, Lake2019HumanFL}---the ability to process a novel sentence and assign it overall meaning(s) by composing the meanings of its individual parts---which, as the name suggests, relies on the compositionality of language itself. A consequence is the ability to comprehend an infinite array of new sentences from only a finite vocabulary~\cite{Chomsky1957, Chomsky2016}.


Whether neural network models can exhibit compositional generalization has been a (sometimes heated) topic of debate for decades, acquiring a somewhat prominent place in the history of AI and cognitive science in the process~\cite{Pollack1990, fodor1993connectionism, Marcus2003,Johnson2004, LakeBaroni2017, Hupkes2019CompositionalityDH}. It is only fairly recently that deep learning methods have been able to deliver unprecedented results in language modeling. However, such performance is provided by extremely large, sample inefficient models. Further, such models do not usually generalize to examples that require the use of language compositionality~\cite{Hupkes2019CompositionalityDH,Dziri2023FaithAF}. It has been argued that the inability of neural networks to leverage compositionality limits them from true natural language understanding~\cite{West2023TheGA,Donatelli2023,Marcus2023ASI}. It is thus natural that we would want neural network models to exhibit compositional generalization, bringing them closer to \emph{true understanding} than mostly \emph{memorization}.

Recent work has started to focus on the idea of building compositional generalization into neural networks using ideas from symmetry, more specifically, using group equivariance \cite{gordon2019permutation, white2022equivariant, QingLi2022, akyurek2023lexsym}. 

We frame our paper as a position piece, arguing that the notion of group equivariance is not sufficient for compositional generalization in language; we must examine symmetry notions beyond the case of actions by groups. More specifically, in the direction of  ``imposing grammar-like rules'' for transduction, which we call \textbf{Generalized Grammar Rules (GGR)}. We elucidate our position by considering the task of transduction between (formal) languages, such that a formal notion of ``meaning'' remains invariant. We propose a general framework where generalized grammar rules serve the same role vis-\`a-vis symmetry in transduction as group actions do in equivariant neural networks~\cite{Cohen2016GroupEC}. We also show that many works in language compositionality already exploit this underlying notion in restricted settings. We argue that formalizing such generalized notions of symmetry will permit a more structured and principled framework to facilitate building neural networks that exhibit better compositional generalization. The situation is analogous to group equivariant networks~\cite{Cohen2016GroupEC}, where presenting the right language and framework~\cite{KondorTrivedi2018, Bekkers20} led to a noticeable acceleration in the development of high performance networks. Further, as \cite{akyurek2023lexsym} point out, formalization of general notions of symmetries will also give a better handle on principled augmentation procedures (via automata rather than symmetry groups).

However, unlike in the case of group equivariance, we cannot usually assume that the underlying symmetries in the language case are known, because outside a few specific tasks such as formal language translation or toy language transduction, rules are not explicit and may have many exceptions. Thus, we propose an error quantification for specific grammar rules, akin to approximate equivariance error quantification in classical equivariance~\cite{Petrache2023, Huang2023ApproximatelyEG}. We present our position under the hypothesis that such an underlying symmetry can be approximated and that a general model for language transducers are automata-like transduction models.

While ``learning the underlying grammar rule equivariance" (learning GGRs) is computationally demanding, we present ideas on how it could be learnt in principle using reinforcement learning (RL) via Markov decision processes (MDPs) and generalizations. In this sense, we show that incorporating symmetry in grammar-based natural language tasks faces the same computational challenges as general RL tasks, indicating strong links between learning the underlying automata equivariance, and RL theory. However, we also note that while automata inhabit a large universe, natural language grammar is much more restricted~\cite{Chomsky1957, Montague1970}, even if it may not seem so at first sight. "Learning automata-symmetries" applied in NLP will allow to detect a hierarchy of symmetries, that can be used to structure more complex and realistic language compositionality/systematicity attempts in the future.

\subsection{Related Works}
Several groups have recently probed the ability of existing neural networks to exhibit compositional generalization. Relevant works include \cite{Johnson2016CLEVRAD,bahdanau2018a,loula2018,deVries2019}. The majority of such works \cite{LakeBaroni2017, gordon2019permutation, Hewitt2020, hupkes2020a, Dankers2022, WhiteCotterel2021, white2022equivariant} have focused on analyzing sequence-to-sequence model performance on toy datasets. In particular, \cite{LakeBaroni2017} (followed by \cite{Ruis2020}), presented a dataset with Simplified version of the CommAI Navigation (SCAN) tasks. While the dataset has proven to be influential for benchmarking compositional generalization of sequence-to-sequence models, \cite{Bastings2018} has shown that an off-the-shelf model can perform perfectly on SCAN without compositional generalization. \cite{Bastings2018} also proposed a complementary dataset to remedy issues with SCAN. On the other hand, \cite{Valvoda2022} argued against using toy datasets, instead proposing the use of artificial string-to-string datasets, generated by sampling subsequential finite-state transducers (SFSTs). The contention is that by controlling the formal properties of SFSTs used to generate the datasets, one could get a better measure of compositional generalization.

In an influential work \cite{gordon2019permutation} considered the hypothesis that \emph{language compositionality can be understood as a form of group equivariance}, constructing a group-equivariant network exploiting local permutation symmetries that could solve most SCAN tasks. \cite{white2022equivariant} go a step further, and build a more expressive group-equivariant model that also
incorporates a group-invariant hard alignment
mechanism. Both of these works acknowledge the limitations of using group equivariance in SCAN-like tasks and conjecture that exploiting more general symmetries will bring about further gains in real world NLP scenarios. A similar message is echoed by \cite{akyurek2023lexsym} who use a compositional adaptation of group symmetries, as a principled mechanism for data augmentation. 

Additionally, there are several works that take the route of program induction, meta-learning etc. for compositional generalization~\cite{nye2020learning,Qiu2021ImprovingCG, Akyrek2020LearningTR, Conklin2021MetaLearningTC,Chen2020CompositionalGV,Csords2021TheDI}, without highlighting the analogy to group-equivariance. It must be noted, however, that grammatical inference \cite{angluin1983inductive} is known to have no upper complexity bound \cite{daley1986complexity}, thus we need to restrict to realistic scenarios.

Motivated by the message of \cite{gordon2019permutation, white2022equivariant, QingLi2022, akyurek2023lexsym}, as well as the successes of incorporating symmetry in computer vision and the physical sciences via group equivariant networks, we present this work as an invitation to extend the concept of symmetries to language in a unified framework.

\section{Transduction between Languages}

A basic notion in NLP is that of \textbf{transduction} which, at a high level, involves two (formal or natural) languages $\mathcal{L}_1$ and $\mathcal{L}_2$ and consists of learning a mapping $\mathcal T: \mathcal{L}_1 \to \mathcal{L}_2$ that preserves some property of interest, such as meaning. 
Transductive NLP tasks are very diverse, including transliteration, translation, speech recognition, spelling correction, temporal phrase normalization, prediction of secondary protein structure from an input sequence etc. An intimately related notion is that of a \textbf{finite-state transducer}, which is an abstract automaton that takes an input string and instead of simply accepting it, converts it into an output string according to some prescribed rules. Before discussing transducers more formally, we discuss more basic notions pertaining to formal languages and Turing machines. 

\subsection{Turing Machines and Finite State Automata}
Recall that a \textbf{Turing Machine (TM)} is defined by a 6-tuple $\mathcal{M} = (S,\Sigma,\Gamma,s_0,H,\delta)$, with the following requirements. $S$ is a finite set of \textbf{internal states}; $\Sigma$ is a finite \textbf{input alphabet}, not containing the pointer symbol $\vartriangleright$, nor the blank symbol $\square$, nor left/right shift symbols $\leftarrow,\rightarrow$; $\Gamma$ is a finite \textbf{tape (or output) alphabet}, not containing $\leftarrow$ and $\rightarrow$, and such that $\Sigma \cup \{\vartriangleright, \square\} \subseteq \Gamma$;  $s_0 \in \Sigma$ is a special state called the \textbf{start state}; $H \subseteq S$ are the \textbf{halting states}; finally, $\delta$ denotes the so-called \textbf{transition function} $\delta:(S \setminus H) \times \Gamma \rightarrow S \times (\Sigma \cup \{\leftarrow, \rightarrow\})$ such that the tape head never erases the $\vartriangleright$ symbol. The transition table of a TM can be encoded in a directed graph over vertex set $S$, which has one edge $(s,s')$ for every choice of $\gamma, \sigma', a \in \Gamma\times(\Sigma\cup\{\leftarrow,\rightarrow\})$ so that $\delta(s,\gamma)=(s',\sigma')$, and in this case the edge $(s,s')$ has label $(\gamma,\sigma')$. Note that \textbf{finite state automata (FSA)} such as deterministic finite automata, are special cases of TMs, being a restricted type of Turing machine without access to a ``scratch'' memory. That is, the tape head can only perform ``read'' operations and always moves from left to right. The ``memory'' that a finite state machine possesses is entirely accounted by what state it is in. Finally, note that the terminology ``finite state automata'' can cause some confusion since TMs can also only have a finite number of states. However, due to the presence of tape memory, TMs can have an infinite number of possible configurations, which is not the case for FSAs. 

\subsection{Transduction between formal languages}
We will mainly concentrate on transduction tasks between \textbf{formal languages}, i.e. languages generated by \textbf{formal grammars}, which by definition are finite collections of production rules. In particular, we are most interested in the class of languages generated by the so-called \textbf{context-sensitive (a.k.a. Type-1) grammars}.

Recall that, by definition, a formal language $\mathcal{L}$ is \textbf{recognized} by a Turing machine $\mathcal{M}$ if the strings on which $\mathcal{M}$ halts form $\mathcal{L}$ i.e. $\mathcal{L} = \{x \in \Sigma^\ast | \mathcal{M} \text{ halts when } <x> \text{is given as an input}\}$. Classes of formal grammars across the Chomsky-Sch\"utzenberger hierarchy~\cite{Chomsky1963TheAT} can be defined through classes of TMs guaranteed to recognize any formal language. Such languages include the larger class of Type-0 grammars. Languages not definable using TMs are \textbf{uncomputable}. There exist interpretable and well understood bijections as below,whose restrictions to classes from the Chomsky hierarchy filtration of formal grammars are well characterized~\cite{HopcroftU79}:
\begin{eqnarray*}
       \{\text{ Turing machines }\}&\stackrel{\text{accepts}}{\longrightarrow}&\{\text{ Formal Languages }\},\\
       \{\text{ Formal Languages }\} &\stackrel{\text{generates}}{\longleftarrow}&\{\text{ Formal grammars }\}.
\end{eqnarray*}
\textbf{Formal transducers} could be thought of as more flexible extensions of TMs in which the condition $\Sigma\subseteq \Gamma$ is relaxed. We can still represent transducers by graphs as in the case of TMs as discussed earlier, but with the labels of edges enriched via output alphabet symbols. If $\mathcal T$ is a transducer with input/output languages $\mathcal L_1,\mathcal L_2$, the associated transduction function can still be denoted by $\mathcal T:\mathcal L_1\to\mathcal L_2$, with a slight abuse of notation.


As stated earlier, transduction involves a mapping between two languages which preserves a property of interest such as meaning. One way of formalizing such properties is via Semantics. Traditional semantics \cite{carnap1959introduction} considers predicates $P(\vec x)$ depending on a set of variables $\vec x$ whose "interpretations" are mappings $\mathcal I(\vec x)$ of the variables to counterparts in world models, and the \textbf{meaning} of $P(\vec x)$ is defined as \emph{the set of interpretations $\mathcal I(\vec x)$ for which $P(\vec x)$ holds true}. 
In a "dual" view, more suitable to our presentation (and with comparable expressive power), "meaning" can be formalized as the assignment $\mathcal L\to \mathcal L_M$ of \textbf{semantic values (SV)} to sentences from a formal language $\mathcal L$. In a universalist view $\mathcal L_M$ would encode all sets of possible interpretations, but in more concrete cases it can boil down to properties relevant to a task such as e.g. syntactic properties of a sentence. It is natural to consider that the semantic value language $\mathcal L_M$ also has a formal language structure, linked to that of the source language $\mathcal L$ via a transducer. 

When introducing compositional symmetries later, the idea that "transduction symmetries preserve meaning" (or semantic values) is an important principle, even if this does not appear in the implementation. This is why we spend some time to formalize the above notion further.

\textbf{Grammars over semantic values.} As mentioned above, we start by postulating the existence of a "grammar of possible meanings" $G_M$, generating language $\mathcal L_M$, whose alphabet $\Sigma_M$ corresponds to "atomic" SV values. Then, a transduction task between languages $\mathcal L_1$ and $\mathcal L_2$ can be thought of as accompanied by "interpretation" (or semantic assignment) transducers $\mathcal I_{1,M}, \mathcal I_{2,M}$ with input languages $\mathcal L_1, \mathcal L_2$, and output language $\mathcal L_M$. Such a transducer \emph{by definition} assigns SVs to sentences from $\mathcal L_i$. 
Then a transducer $\mathcal T_{12}:L_1\to L_2$ \textbf{"preserves meaning"} if $\mathcal T_{12}\circ \mathcal I_{2,M}=\mathcal I_{1,M}$. This means that for each sentence from $\mathcal L_1$ if we assign it an interpretation by $\mathcal I_{1,M}$ or first apply $\mathcal T_{12}$ and then assign the result an interpretation by $\mathcal I_{2,M}$, we get the same result.
\begin{rmk}
    "Universal" interpretation functions $\mathcal I_{i,M}$ are usually not explicitly known, and hard to encode usefully for realistic scenarios. Thus, it must be emphasized that the described setup does not postulate the absolutist view that "there exists a universal meaning", instead we take a relativist/pragmatist stand, and consider ad-hoc versions $\mathcal L_M, \mathcal I_{i,M}$ for each given scenario. 
\end{rmk}

\section{Symmetries in Language Transduction}

In the previous section, we discussed transduction between formal languages, along with a formal notion of meaning, which stays invariant during the transductive process. In this section, we use this \emph{meaning invariance} as a starting off point to describe general ``language symmetries,'' while also comparing them to symmetries discussed in classical group action contexts, which is the bread and butter of equivariant networks~\cite{Cohen2016GroupEC, KondorTrivedi2018}. As the reader is more likely to be familiar with the classical case, we discuss it first.

\subsection{Basic framework for classical Equivariance}
For the sake of completeness, we begin by recalling that a \textbf{group} is a set $G$, endowed with a binary operation $G\times G\to G$ satisfying the associative property, and such that an identity $e\in G$ exists, and such that all elements $g\in G$ have an inverse in $G$. 

\begin{rmk}
    In order to define equivariance, we only require a binary operation, and not the remaining requirements from the definition of a group.
\end{rmk}

\textbf{Equivariance.} Consider two spaces $X, Y$ and a function $f:X\to Y$. For defining equivariance we need to fix \textbf{$G$-actions} on $X$ and $Y$, i.e. maps $\rho_X:G\to \mathcal T_X, \rho_Y:G\to\mathcal T_Y$ with $\mathcal T_X\subseteq\{T:X\to X\}, \mathcal T_Y\subseteq \{T:Y\to Y\}$ such that for all $g,h\in G$, the following holds
\begin{equation}\label{eq:groupact}
    \rho_X(gh) = \rho_X(g)\circ\rho_X(h), \quad \rho_Y(gh)=\rho_Y(g)\circ \rho_Y(h).
\end{equation}

Assume $G$ is a space with a binary operation $G\times G\to G$. Given a function $f:X\to Y$ and actions $\rho_X,\rho_Y$ as above, we say that \textbf{$f$ is $G$-equivariant} if for all $g\in G$ we have
\begin{equation}\label{eq:equivar}
f\circ(\rho_X(g)) = (\rho_Y(g))\circ f.
\end{equation}
A particular case is that $\rho_Y(g)$ is the identity of $Y$ for all $g\in G$, in which case \textbf{$f$ is $G$-invariant}: $f\circ (\rho_X(g))= f$ for all $g\in G$.

The notion of equivariance has led to the development of equivariant neural networks, a fruitful area of research within machine learning, which has been especially successful in applications in the physical sciences. The main benefits of equivariant neural networks stem from the fact that they encode a clear symmetry of maps from $\mathcal T_X, \mathcal T_Y$ \emph{explicitly} into the model layer structure. Thus, they enable encoding label functions more efficiently. Crucially, eqs. \eqref{eq:groupact}, \eqref{eq:equivar} are thus not the main reason of interest in equivariance, as is the fact that structural results from group representation theory and the algebra of convolutions, furnish a toolbox for constructing maps from $\mathcal T_X, \mathcal T_Y$ efficiently.



\subsection{Generalized Grammar Rules as Symmetries in Transduction}

The notion of "language symmetries" that we propose has the following crucial property in common with group-equivariance symmetries: The structure of the underlying task permits one to encode, in a short and formally defined form, a large set of transformations that preserve the truth values of the task being investigated, which in turn are interpretable as structural properties of the task. We propose two ways of formalizing language symmetries: Using quotient operations over automata, and using predicates over grammars. While we describe both, it is the latter that is operationally more relevant. 

\subsubsection{Quotient operations on Turing Machines}
Let $\mathcal T=(S, \Sigma, \Gamma, s_0,H,\delta)$ be a (finite) transducer. If $\sim$ denotes an equivalence relation over the set of states $S$, then we may define the corresponding \textbf{quotient transducer} 
\[\mathcal T/\sim=(S/\sim,\Sigma,\Gamma,[s_0], \overline H, \overline \delta)
\]
where the new internal state space $S/\sim$ comprises the set of $\sim$-equivalence classes from $S$, $\overline H$ are equivalence classes that contain an element of $H$, and the transition table $\overline \delta$ is defined by setting $\overline\delta([s], \gamma) = ([s'],\sigma')$ whenever there exists $s\in[s], s'\in [s']$ such that $\delta(s,\gamma)=(s',\sigma')$. This can be also obtained via a suitable quotient of the underlying computational graph, which identifies nodes according to the quotient $S\mapsto S/\sim$ and defines halting states $\overline H$.

\textbf{Quotient-symmetric TMs.} In case we add the restriction $\Sigma\subset \Gamma$, we obtain a definition of \textbf{quotient TM's} as a special case. Note that if a TM $\mathcal M$ recognizes $L(\mathcal M)$, then in general, a quotient of it $\mathcal M'=\mathcal M/\sim$ will \emph{recognize a language equal to, or larger than} that recognized by $\mathcal M$ i.e. $L(\mathcal M')\supseteq L(\mathcal M)$, as a consequence of collapsing state sequences. We say that $\mathcal L=L(\mathcal M)$ has \textbf{symmetry induced by $\sim$ quotient} (or \textbf{$\sim$ symmetric}) if $L(\mathcal M/\sim)=L(\mathcal M)$. In this case, $\mathcal M/\sim$ is a "simpler TM equivalent to $\mathcal M$".

If by abuse of notation we think of $\mathcal M$ as a function $\mathcal M:\Sigma^*\to\{0,1\}$ which assigns "1" to accepted strings and "0" to the others, then the passage to the quotient $\mathcal M\mapsto\mathcal M/\sim$ is a Turing Machine analogue to the case of improving a group-invariant classification task with labels $\{0,1\}$ by replacing a model by an group-invariant model with fewer parameters.

%

The extension to transducers, of which TMs are a special case, is direct: we say that trasducer $\mathcal T$ is \textbf{$\sim $ symmetric} if $\mathcal T/\sim$ induces the same map $\Sigma^*\mapsto \Lambda^*$ as $\mathcal T$. In the analogy to group-equivariance, quotient transducers correspond to passing to a suitable quotient in group-equivariant tasks.


Quotient symmetries are natural and parallel familiar group-equivariant setups, however it seems hard to make them practically useful, as a good understanding/discussion of specific transducer dynamics seems necessary for implementing a useful quotient symmetry. We present a more direct approach in the next section.

\subsubsection{Generalized grammar rules}
Perhaps the most straightforward way for fixing symmetries in language transduction is in terms of predicates over (transduction) grammars, i.e. rules for transducers $\mathcal T$ with input/output languages $\mathcal L_1, \mathcal L_2$ with alphabets $\Sigma, \Lambda$. Then a \textbf{Generalized Transduction Rule (GTR)} for $\mathcal T$ is a predicate of the form:
\begin{multline}\label{eq:gtr}
    \forall 1\le i\le h_1, \forall 1\le j\le h_2, \forall x_i\in C_i, \forall y_j\in D_j, \\
    B_0\mathcal T(A_1)B_1\cdots \mathcal T(A_{k_1})B_{k_1}=\\\overline B_0\mathcal T(\overline A_1)\overline B_1\cdots\mathcal T(\overline A_{k_2})\overline B_{k_2}, 
\end{multline}
in which $h_1,h_2,k_1,k_2$ are nonnegative integers with $k_1>0$ and we have used:
\begin{itemize}
    \item two finite sets of \emph{symbols for variable strings} $X=\{x_1,\dots, x_{h_1}\}, Y=\{y_1,\dots,y_{h_2}\}$, and a corresponding set of subclasses of the input/output languages $C_1,\dots,C_{h_1}\subseteq L_1$ and $D_1,\dots,D_{h_2}\subseteq L_2$;
    \item two collections of strings $A_1,\dots, A_{k_1}, \overline A_1,\dots,\overline A_{k_2}\in (X\cup \Sigma)^*$ and two collections of strings $B_0,\dots,B_{k_1},\overline B_0,\dots,\overline B_{k_2}\in(Y\cup \Lambda)^*$, such that for each $1\le i\le h_1$ the symbol $x_i$ appears in at least one of the $A_j, \overline A_j$, and for each $1\le i\le h_2$ the symbol $y_i$ appears in at least one of the $B_j, \overline B_j$.
\end{itemize}

\begin{ex}
To make the above more concrete, consider two examples of GTRs. For $a_i\in\Sigma^*, b_i\in\Lambda^*$:
\begin{multline*}
\forall x_1,x_2\in \Sigma^*, T(x_1 a_1 a_2 x_2 a_3) = \\ b_1 T(x_1) T(x_1) T(a_1 a_3) T(x_2a_2) b_2.
\end{multline*}
\vspace{-7mm}
\begin{multline*}
\forall x_1,x_2\in \Sigma^*, \forall y_1\in \Lambda^*, T(x_1 a_1 a_2)b_1y_1T(x_2 a_3) = \\ b_1 T(x_1) T(x_1) T(a_1 a_3) y_1 T(a_1x_2a_2) b_3y_1.
\end{multline*}
\end{ex}
While our treatment could in principle apply for GTRs, we will actually restrict to a special case. We say that a GTR of the form \eqref{eq:gtr} is a \textbf{Generalized Grammar Rule (GGR)} if
\begin{itemize}
    \item it involves no variable $y_j$ over the output language,
    \item we have $k_1=1$ and the left hand side is composed only of $\mathcal T(A_1)$;
    \item the length of each $\overline A_i$ is shorter or equal to the length of $A_1$.
\end{itemize}
Thus the form of a GGR is (taking modifications of \eqref{eq:gtr} as above, and notation changes $h=h_1$, $A=A_1$, $k_2=k$)
\begin{multline}\label{eq:rule}
    (\mathcal R):\ \forall 1\le i\le h, \forall x_i\in C_i, \\
    \mathcal T(A)=\overline B_0\mathcal T(\overline A_1)\overline B_1\cdots\mathcal T(\overline A_k)\overline B_k.
\end{multline}
Our first example was a GGR while the second was not.

If a GTR or a GGR holds for a transducer $\mathcal T$, this constrains the transducer behavior, and at the same time, due to the universal quantifiers in \eqref{eq:gtr}, it encodes information of possibly extensive sets of relations about the graph of the transduction function induced by $\mathcal T$. GGRs play a special role, because they can be considered as production rules for transducer $\mathcal T$, i.e. they prescribe how to transduce a string of the form $A_1$, once strings of shorter length have been transduced. Thus GGRs are directly suitable for data augmentation, whereas general GTRs are not.

We now introduce an error function that measures "how far from holding" for a given transducer $\mathcal T$ a given GGR is. With the same notation as in \eqref{eq:rule}, for a fixed value of "temperature" parameter $\beta>0$ we set 
\begin{multline}\label{eq:errorggr}
    \mathsf{Err}_\beta(\mathcal T, \mathcal R):=\\\sum_{\substack{a_i\in C_i\\1\le i\le h}}\exp\big((-\beta - \log(\sharp\Sigma))\sum_{j=1}^h\ell(a_j)\big) \mathsf{dist}(\widehat A, \widehat B),
\end{multline}
in which $\ell(\alpha)$ is the length of a string $\alpha$, and $\mathsf{dist}(\alpha_1,\alpha_2)$ is the \emph{Levenshtein distance} between strings $\alpha_1,\alpha_2$ (we may replace it by a task-dependent combinatorial distance in applications), $\widehat A\in \Sigma^+$ is the string obtained by replacing $x_i=a_i$ for $1\le i\le h$ in $A$, and $\widehat B\in \Lambda^*$ is obtained by the same replacement in the right hand side of \eqref{eq:rule}, i.e. in $\overline B_0\mathcal T(\overline A_1)\cdots \overline B_k$.

The factor $\exp(\cdots)$ in \eqref{eq:errorggr} is a natural normalization, which is useful in order to have the finitenes guarantee $\mathsf{Err}_\beta(\mathcal T,\mathcal R)$ proved in below Prop. \ref{prop:length}. 
This result, valid even if the classes $C_i, D_i$ contain infinite sequences of strings of unbounded length, works under the following mild assumption on $\mathcal T$: we say that $\mathcal T$ has \textbf{at most $D$-power growth}\footnote{By inspecting the proof of Prop. \ref{prop:length}, it can be verified that if $\mathcal T$ has higher than polynomial growth and $\log(\ell(\mathcal T(A))/\log(\ell(A))\leq F(\ell(A))$ for some $F:\mathbb N\to[0,+\infty)$, then a variant of the result may still hold, if we replace normalization factor by $\exp\left((-\beta F(\ell(A))  -\log(\sharp\Sigma))\sum_j\ell(a_j)\right)$. However this seems like a technical point and is not developed further in this setting.} 
if there exist $C>0$ and $D\in \mathbb N$ such that for all input strings $\alpha\in L_1$ there holds $\ell(\mathcal T(\alpha))\leq C (\ell(\alpha))^D$.

\begin{prp}[Proved in App. \ref{app}]\label{prop:length}
    Assume that $\mathcal T$ has $D$-power growth. Let $(\mathcal R)$ be a GRR with notation as in \eqref{eq:rule}. Then there exists a constant $C$ depending only on the growth bound for $\mathcal T$ and on $h, A, \overline B_0,\dots,\overline B_k, \overline A_1,\dots,\overline A_k$ and $\beta>0$, such that $\mathsf{Err}_\beta(\mathcal T, \mathcal R)\leq C$.
\end{prp}
The idea of the proof is that $\mathsf{dist}(\widehat A, \widehat B)$ only grows polynomially in the lengths $\ell(a_i)$, whereas the factor $\mathsf{exp}\left((-\beta - \log(\sharp\Sigma))\sum_j \ell(a_j)\right)$ decays exponentially, allowing the convergence of the sum. The reason why the above convergence guarantee is important, is that only having a finite value of the error will allow to set up gradient descent procedures based on this error function, for "fitting a set of GGRs to a given transducer" or "learning approximate GGRs underlying the data".

\begin{rmk}\label{rmk:simplerrules}As a consequence of the form of \eqref{eq:errorggr}, note also that simpler rules have lower length growth of $\hat A, \hat B$ (and thus of $\mathsf{dist}(\widehat A,\widehat B)$) in terms of the $\ell(a_i)$, and as a consequence, in a setup where we learn symmetries, \textbf{learning simpler rules first} is favoured by our error function, as the exponential decay will "kick in" at lower values of $\ell(a_i)$, the more complicated our rules.
\end{rmk}

\section{Examples Uses of Language Symmetries in Transduction}

In the previous sections, we elucidated a general notion of language symmetry via GGRs and connected it to the notion of classical equivariance. In this section, we show that several recent works that exploit similar ideas, in fact can be recovered as special cases of our treatment, also serving to further clarify it.

\subsection{String Equivariance of \cite{gordon2019permutation}}
\begin{ex}[String equivariance]\label{ex:gordon1}
    Gordon et al. \cite{gordon2019permutation} is one of the first and most influential works  for the use of equivariance towards compositional generalization in natural language processing contexts. In particular, they work with a "local" permutation action $T_g:\Sigma\to\Sigma, g\in G$, such that an extension to $\Sigma^*\to\Sigma^*$ can be defined by
\begin{multline}\label{gordonrule}
    \forall k\forall x_1,\dots,x_k\in \Sigma,\ T_g(x_1\dots x_k)=\\
    T_g(x_1)\dots T_g(x_k)
\end{multline}
based on this, an implementation of an equivariant sequence to sequence model based on group convolutions is proposed. 

In our setting, given a set of $T_g, g\in G$ as above, we can get the GGRs (one for each element $g\in G$) defined as
\[
    (R_g): \forall x_1,x_2\in \Sigma^*,\quad T_g(x_1x_2)\to T_g(x_1)T_g(x_2),
\]
which permit the generation of all the rules in \eqref{gordonrule} by composition.
\end{ex}
\begin{ex}[Context sensitivity]\label{ex:gordon2}
    Another scenario described in \cite{gordon2019permutation} but left as an open direction is that of conjunction replacement. The task involves, for example, rules allowing to encode the following:
    \begin{eqnarray*}
    T(\text{RUN LEFT AND WALK}) &=& \text{LTURN RUN WALK}, \\
    T(\text{RUN LEFT AFTER WALK}) &=& \text{WALK LTURN RUN}.
    \end{eqnarray*}
\cite{gordon2019permutation} note that for this task, compositional generalization can not be achieved via their framework of local equivariance. To see this, notice that changes induced by the replacement of AND by AFTER in the input language can not be implemented by any local group operation in the output language.    
    
While permutation symmetries are not able to express such rules, in our setting we can produce rules that can. For instance, we can express the difference between "AND" and "AFTER" as:
    \begin{eqnarray*}
    \forall x_1,x_2\in \Sigma^*,\quad T(x_1\text{ AND }x_2) &\to& T(x_1) T(x_2),\\ \forall x_1,x_2\in \Sigma^*,\quad T(x_1\text{ AFTER }x_2) &\to& T(x_2) T(x_1).
    \end{eqnarray*}
    More generally, these rules can encode \emph{context sensitive transduction} rules, which are thus lexical symmetries.
\end{ex}

\subsection{Equivariant Transduction via Invariant Alignment of \cite{white2022equivariant}}

We will show here how we can use GGRs to extend the setting of \cite{white2022equivariant}.

\begin{ex}[Grammar tagging and annotations]\label{ex:white1}
     The work of \cite{white2022equivariant} generalizes \cite{gordon2019permutation} in the following sense: The permutations $T_g$ are applied but \emph{conditioned over lexical classes}. In other words, symmetries are only allowed to exchange words inside a given fixed lexical class (examples of lexical classes are subsets of verbs/adjectives/nouns/etc. of interchangeable semantic roles), after which the extension to $\Sigma^*$ is performed as shown in Example \ref{ex:gordon1}. This can be achieved by allowing \emph{part-of-speech taggers}. But of course, we can even include conditioning over a wide range of annotations, \emph{beyond the (local) level of single word tagging}, via more general \textbf{grammatical taggers}. We can condition the switch of a whole substring by another, in the presence of a suitable tagging of the former. In order to implement conditioning on annotations of $N$ possible classes, we formalize annotations, i.e. extend $\Sigma$ via new letters $\{l_i, l_i':\ i=1\dots, N\}$ (to be thought as class-dependent brackets), and assume that the input sentences include annotations, in which we interpret substring $l_i A l_i'$ as the annotation that $A$ is in grammatical class $i$.
    Then, in order to apply $T_g$ to exchange $x$ with $y=T_g(x)$ only restricted to substrings from grammatical class $i$, we include the rules $\forall x\in \Sigma^*,\quad T_g(l_i x l_i')=l_i y l_i$. 
    Notice that this is only imposing $T_g$-action to strings tagged by $l_i,l_i'$ delimiters.
\end{ex}

\subsection{Toy Transducer Rules in \cite{lake2023human}}
\begin{ex}[Toy transducer rules are GGRs]\label{ex:lakebaroni}
    In \cite{lake2023human}, toy model languages were generated in order to test systematicity, which can be thought of as a weakening of compositionality. Notably, systematic generalization has remained an open challenge in neural networks for over three decades, and has been the subject of a long standing debate. \cite{lake2023human} showed that standard neural network models, if optimized for their compositional skills, can do a reasonable job at mimicking human systematic generalization. Interestingly, \cite{lake2023human} tested for systematicity via \emph{interpretation grammars}, which in fact are GGRs. In their setup, the input alphabet $\Sigma$ consisted of a finite set of words such as "zup", "fep", "tufa", etc. and the output alphabet $\Lambda$ was a finite set of colors. An example of interpretation grammar is
    \begin{eqnarray*}
        T(\text{zup}) &=& [green],\\
        T(\text{fep}) &=& [rose],\\
        T(\text{gazzer})&=&[red],\\
        T(\text{tufa})&=&[bourbon],
    \end{eqnarray*}
    \vspace{-9mm}
    \begin{multline*}\forall x_1, x_2\in \Sigma,\ T(x_1\text{ lug }x_2)=\\T(x_2)T(x_1)T(x_2)T(x_1)T(x_1),
    \end{multline*}
    \vspace{-8mm}
    \begin{eqnarray*}
        \forall x_1,x_2\in \Sigma,\ T(x_1\text{ kiki }x_2)&=&T(x_1)T(x_2),\\
        \forall x_1\in \Sigma,\ T(x_1\text{ blicket})&=&T(x_1)T(x_1),\\
        \forall x_1,x_2\in \Sigma,\ T(x_1x_2)&=&T(x_1)T(x_2).
    \end{eqnarray*}
    Each line above is a GGR as in our setup, but with the restriction that left hand sides of all the equations have the same form $T(\alpha)$ in \cite{lake2023human}. There is yet another difference that $x_1,x_2$ may vary over elements of $\Sigma$ rather than over $\Sigma^*$ as in our setting. The translation to our setting can be done by using a single tagger for words, and replacing words $w$ by $lwl'$ on the left of the quantified equations. Thus rules 1-4 stay the same while e.g. the 5th rule above becomes
    \begin{multline*}
    \forall x_1, x_2 \in \Sigma^+,\ T(lx_1l' l\text{ lug }l'lx_2l)= \\ T(x_1)T(u_1)T(x_1)T(u_1)T(u_1).
    \end{multline*}
    Then we may stipulate that the input language $\mathcal L_1$ comes with suitable well-formed tagging, fitting the above setup.
\end{ex}

\section{Learning Transduction Symmetries}

In the previous section we showed that several recent works on language compositionality could be seen as special cases of our framework. Here, we consider the questions of ``learning symmetries'' in transduction, in the sense of ``learning transducers with approximate\footnote{Note that allowing \textbf{approximate} symmetries has always been part of the definitions of compositionality: e.g. \cite{fodor1993connectionism} says "insofar as a language is systematic, a lexical item must make \emph{approximately} the same semantic contribution to each expression in which it occurs" (with emphasis added here).} symmetry constraints''. Presenting a full approach to symmetry learning is a challenging open problem even for classical (group) equivariance. Therefore, our goal is not to propose a solution, but rather to point out the relevance of these questions within our framework. We expect that this will help orient the reader and, as we will see, it will permit reinterpretations of several important open questions.

\subsection{Learning approximate symmetries: motivation and connection with Reinforcement Learning}

How do we infer what symmetries are the best to impose on a transducer for a given task? Beyond certain applications, such as transduction between programming languages (say from a high-resource language such as Python, to a low-resource one such as Verilog), or grammars pertaining to toy models, as the transduction tasks become harder, the symmetries are not explicitly known. We may have access to only a handful of generalized grammar rules, whereas the remaining ones have to be learned. The finiteness result of Prop. \ref{prop:length}, allows to formulate approximate GGRs and perform optimization over $\mathsf{Err}_\beta$ or related loss functions, allowing to learn GGRs adapted to a task.

Previous attempts towards this task include ``neural program interpreters'' \cite{reed2015neural} and ``neural turing machines'' \cite{zaremba2015reinforcement}, which fit within the purview of  ``program induction''. For more recent examples, neural network models such as transformers \cite{strobl2023transformers} or RNNs \cite{orvieto2023resurrecting} have been evaluated on tasks formulated as formal language recognition. 

Putative GGRs $\mathcal R$ over a transduction task $\mathcal T$ can be encoded as a not completely determined transducer, given as
\[
\mathcal T_{\mathcal R}=(S_{\mathcal R}, \Sigma\cup X_{\mathcal R}, \Gamma\cup X_{\mathcal R}, s_0,H,\delta_{\mathcal R}).
\]
A simple way to formalize learning $\mathcal R$ is to think of $S_{\mathcal R}, X_{\mathcal R}, \delta_{\mathcal R}$ probabilistically: we consider $\mathcal T_{\mathcal R}$'s computational graph as a random graph whose edges it is our task to learn. In practice we may fix $S_{\mathcal R}, X_{\mathcal R}$, and determine $\delta_{\mathcal R}$'s distribution from data. This setting is akin to Markov Decision Processes (MDP) from classical Reinforcement Learning theory (e.g. as appearing in \cite{abate2022learning}).

Note that in this formalization, we would allow errors in GGR fitness to the model, in order to formulate a tractable differentiable learning task. On the other hand, similar to the case of approximate equivariance, allowing symmetry errors in prescribed-symmetry models, may also help to allow for exceptions in real-world models and to make the models more robust. This consideration directly translates to our proposed transducer symmetries, and can be implemented via the  above "probabilistic transducers" framework.

The novelty of our main position applied here, besides the superficial change of working with transducers rather than the classical RL setting, is that we \textbf{reinterpret the RL theory as a learning task for transducer symmetries}. Beyond the already explored Equivariance - GGR link, Some consequences of this viewpoint come from the link from GGRs and Equivariance, to RL are as follows: 
\begin{itemize}[noitemsep]
        \item Consequence 1: a geometric view of transduction and language processing, saying that e.g. \emph{transduction tasks are like exploration by an agent, but in language space}, justifies that spatial RL strategies should be relevant in transduction tasks.
        \item Consequence 2: using the bridge from RL to equivariance theory, we can state that \emph{learning (group) symmetries could benefit from RL approaches}. 
        \item Consequence 3: \emph{RL can be thought of as a transduction task with symmetries}: the link from RL to transduction is well accepted, and the parallel to equivariance theory indicates that our framework may allow e.g. to compare numbers of degrees of freedom under different RL structures as in \cite{Petrache2023}.
        \end{itemize}
\subsection{Further examples of links to conceptual themes in language processing} 
There are several directions that seem closely related and may benefit from the parallel to learning transducer symmetries, in conjunction to the RL setup sketched above. 

\textbf{Structural form search.} A first direction is work on ``discovery of structural form'' \cite{kemp2008discovery}, in which a framework for building graph grammars underlying data is indicated. Applying the above RL approach to automata search can be of interest for searching amongst geometric forms underlying data. 

\textbf{Language based on agent models.} \cite{andreas2022language} proposes to treat language models as "agent models" in which interpretation is based on modelling of an agent's underlying internal states. In our framework, we may directly formulate these internal states at the level of the transducers learning the task, and learning can proceed as an RL agent, adapting its GGR to the underlying task. Learnable hidden internal states then play a similar role as learnable GGR constraints. Note that this formalization is quite different from the one of \cite{andreas2022language}, although the underlying task interpretation coincides.

\textbf{Words as cues.} It has been argued \cite{lupyan2019words} that interpreting human lexical processing in terms of tokens and semantic interpretation can run into problems, which become evident when treating transduction between unrelated (natural) languages. A "words-as-cues" view then proposes a less structured approach, in which language/semantics just give inductive biases on the interpretations of multimodal inputs. This view can be construed as conceptually similar to ours, in which GGR errors could be quantifying these biases. We emphasize the parallel to classical equivariance, in which the source of inductive bias were more classical/physical symmetries of the tasks.

\textbf{Reformulating binding and disentanglement.} The implementation of GGRs allows attacking the so-called \emph{binding problem} (see \cite{greff2020binding}) in transduction and related tasks, from the point of view of generalized equivariance. This can bring possible benefits: for example, the reinterpretation of GGRs as symmetries allows to extend the definition of disentanglement \cite{higgins2018towards}, initially based on classical equivariance, to transduction problems.

\section{Conclusions}
Based on a theory of meaning modeled via formal languages and transducers, we have proposed a framework in which Generalized Grammar Rules allow to impose compositional generalization constraints on learning tasks, and give a direct method for compositional data augmentation. We have argued that GGRs can be imposed in transduction tasks, bearing parallels to the imposition of group-equivariance in physics-inspired models with natural group actions. This has been done by formulating a natural quantification for the validity error of GGRs, analogous to approximate equivariance error quantification. We then described how a series of recent works related to compositionality in transduction fit within our framework, and then described how our framework can be used in tasks where GGRs can be learned. Here GGR learning can be viewed as the learning of  compositional symmetries, with direct relations to RL strategies, also bearing interesting consequences upon RL theory. Finally, we indicated how our framework connects to adjacent research topics such as structural form search, words-as-cues models, agent models, and in problems related to compositionality, such as binding and disentanglement. 

\section{Potential Broader Impact}
This work is theoretical, presenting a new point of view towards compositionality. The underlying views may have consequences if implemented, however potential impact does not follow directly from our paper.

\nocite{langley00}

\bibliography{example_paper}
\bibliographystyle{icml2024}

\newpage
\appendix
\onecolumn
\section{Proof of Proposition \ref{prop:length}}\label{app}
We refer to the notation of \eqref{eq:rule} and \eqref{eq:gtr} here, and the goal is to study the convergence of the sum from \eqref{eq:errorggr} (also reproduced in \eqref{eq:errorggr1} below, for the reader's convenience). We start with a few auxiliary bounds. First, we bound the number of summands in each term in \eqref{eq:errorggr}.
\begin{equation}\label{eq:numberCi}
    \sharp C_i^{\ell}:=\sharp \left(C_i\cap \left\{x\in L_1:\ \ell(x)= \ell\right\}\right) \leq \sharp\left\{x\in \Sigma^*:\ \ell(x)=\ell\right\}=(\sharp\Sigma)^\ell.
\end{equation}
Next, consider the term $\mathsf{dist}(\widehat A, \widehat B)$. As $\mathsf{dist}$ is Levenshtein distance (equalling the minimum number of insertion/deletions required to pass from $\widehat A$ to $\widehat B$, it follows that 
\begin{equation}\label{eq:levlength}
    \mathsf{dist}(\widehat A, \widehat B) \leq \ell(\widehat A)+\ell(\widehat B).
\end{equation}
We now control from above $\ell(\widehat A), \ell(\widehat B)$. The power-$D$ growth hypothesis on $\mathcal T$ states that for all $\alpha\in L_1$ we have $\ell(\mathcal T(\alpha))\leq C_{\mathcal T} (\ell(\alpha))^D$. For the case of $\ell(\widehat A)$ this implies that, for any $A\in (X\cup \Sigma)^*$, 
\begin{eqnarray}\label{eq:ellA}
    \ell(\widehat A)&=&\ell\left(\mathcal T\left(\left.A\right|_{x_i=a_i, 1\le i\le h}\right)\right)\leq C_{\mathcal T}  \left(\ell\left(\left.A\right|_{x_i=a_i, 1\le i\le h}\right)\right)^D\nonumber\\
    &\leq& C_{\mathcal T} \left(\ell(A) + \sum_{i=1}^h d_i\ell(a_i)\right)^D\leq C_{\mathcal T} \left(\ell(A) +  d_{\max} \ell(\vec a)\right)^D\nonumber\\
    &\leq&C_{\mathcal T} \overline C_\mathcal R(\ell(\vec a) + 1)^D,
\end{eqnarray}
where in the above  
\begin{itemize}
\item $d_i$ denotes the total number of instances of variable $x_i$ in $A$
\item we denote $d_{\max}:=\max_{1\le i\le h}d_i$
\item we denote $\ell(\vec a):=\sum_{j=1}^h\ell(a_j)$.
\end{itemize}
We now proceed to a similar estimate for $\ell(\widehat B)$. Note that $\widehat B$ is obtained from the right hand side of \eqref{eq:rule} by replacing $x_i=a_i$ for all $1\le i\le h$ and by computing $\mathcal T(\overline A_i)$ for all $1\le i\le k$. We may apply a bound similar to \eqref{eq:ellA} to each such term, and we find 
\begin{eqnarray}\label{eq:ellB}
    \ell(\widehat B)&=&\sum_{i=0}^k\ell(\overline B_i) + \sum_{i=1}^k\ell(\mathcal T(\overline A_i))\leq \sum_{i=0}^k\ell(\overline B_i) + C_{\mathcal T} \sum_{i=1}^k\left(\ell(\overline A_i) + d_{\max}\ell(\vec a)\right)^D\nonumber\\
    &\leq&C_{\mathcal T} ( C_{\mathcal R}^{(1)} + \overline C_{\mathcal R})(\ell(\vec a)+1)^D\leq C_{\mathcal T} \widetilde C_{\mathcal R}(\ell(\vec a) + 1)^D. 
\end{eqnarray}
Now we may plug \eqref{eq:ellA}, \eqref{eq:ellB} into \eqref{eq:levlength} and then plug the outcome together with \eqref{eq:numberCi} into \eqref{eq:errorggr} to find the following bounds (the first line is a copy of \eqref{eq:errorggr}, for the convenience of the reader:
\begin{eqnarray}\label{eq:errorbd}
    \mathsf{Err}_\beta(\mathcal T, \mathcal R)&:=&\sum_{a_1\in C_1,\dots,a_h\in C_h}\exp\left((-\beta - \log(\sharp\Sigma))\sum_{j=1}^h\ell(a_j)\right) \mathsf{dist}(\widehat A, \widehat B)\label{eq:errorggr1}\\
    &\leq&C_{\mathcal T}(\widetilde C_{\mathcal R}+\overline C_{\mathcal R})\sum_{a_1\in C_1,\dots,a_h\in C_h}\exp\left((-\beta - \log(\sharp\Sigma))\ell(\vec a)\right)(\ell(\vec a)+1)^D\nonumber\\
    &=&C_{\mathcal T}(\widetilde C_{\mathcal R}+\overline C_{\mathcal R})\sum_{\vec\ell\in \mathbb N^h}\sum_{a_1\in C_1^{\ell_1}}\cdots\sum_{a_h\in C_h^{\ell_h}}\exp\left((-\beta - \log(\sharp\Sigma))\|\vec\ell\|_1\right)(\|\vec\ell\|_1+1)^D\nonumber\\
    &\stackrel{\text{\eqref{eq:numberCi}}}{\leq}&C_{\mathcal T}(\widetilde C_{\mathcal R}+\overline C_{\mathcal R})\sum_{\vec\ell\in \mathbb N^h}(\sharp\Sigma)^{\|\vec\ell\|_1}\exp\left((-\beta - \log(\sharp\Sigma))\|\vec\ell\|_1\right)(\|\vec\ell\|_1+1)^D\nonumber\\
    &=&C_{\mathcal T,\mathcal R}\sum_{\vec\ell\in \mathbb N^h}e^{-\beta \|\vec\ell\|_1}(\|\vec\ell\|_1+1)^D,
\end{eqnarray}
where we denoted $C_{\mathcal T, \mathcal R}:=C_{\mathcal T}(\widetilde C_{\mathcal R} + \overline C_{\mathcal R})$, and the sum in \eqref{eq:errorbd} converges for any $\beta>0$, and is bounded depending only on $h, \beta, D$, in which $h,D$ in turn depend only on $\mathcal R,\mathcal T$.


\end{document}